\documentclass[10pt,journal,compsoc]{IEEEtran}
\usepackage[latin1]{inputenc}
\usepackage{tikz}
\usepackage{graphicx}
\usepackage{relsize}
\usepackage{pdfpages}
\usepackage{epstopdf}
\usepackage{hyperref}
\usepackage{cleveref}
\DeclareGraphicsExtensions{.png,.pdf}

\usepackage{algorithm2e}
\graphicspath{ {./images/} }
\usetikzlibrary{shapes,arrows}

\ifCLASSOPTIONcompsoc
  \usepackage[nocompress]{cite}
\else
  \usepackage{cite}
\fi

\ifCLASSINFOpdf

\else

\fi

\begin{document}

\title{Applying support vector data description for fraud detection \\}

\author{Mohamad~Khedmati,~\IEEEmembership{University of Birjand,}
        Masoud~Erfani,~\IEEEmembership{University of New Brunswick,}
        and~Mohammad~GhasemiGol,~\IEEEmembership{Assistant professor at University of Birjand}
\IEEEcompsocitemizethanks{\IEEEcompsocthanksitem Mohammad GhasemiGol is with the Department
of Electrical and Computer Engineering, University of Birjand,
.\protect\\
E-mail: ghasemigol@birjand.ac.ir
\IEEEcompsocthanksitem Masoud erfani is a student at University of New Brunswick and Mohammad khedmati is a student at University of Birjand,
.\protect\\
\IEEEcompsocthanksitem Masoud erfani email is  merfani@unb.ca and
Mohammad khedmati email is mkhedmati@birjand.ac.ir
.\protect\\
\IEEEcompsocthanksitem Density-based spatial clustering of applications with noise(DBSCAN)- support vector data description(SVDD)-Support-vector machine(SVM)- Chi-square automatic interaction detection(CHAID)
.}
\thanks{
.}}


\IEEEtitleabstractindextext{%
\begin{abstract}
Fraud detection is an important topic that applies to various enterprises such as banking and financial sectors, insurance, government agencies, law enforcement, and more.  Fraud attempts have been risen remarkably in current years, shaping fraud detection an essential topic for research. One of the main challenges in fraud detection is acquiring fraud samples which is a complex and challenging task. In order to deal with this challenge, we apply one class classification methods such as SVDD which does not need the fraud samples for training. Also, we present our algorithm REDBSCAN which is an extension of DBSCAN to reduce the number of samples and select those that keep the shape of data.
The results obtained by the implementation of the proposed method indicated that the fraud detection process was improved in both performance and speed.
\end{abstract}

}

\maketitle

\IEEEdisplaynontitleabstractindextext

%
\IEEEpeerreviewmaketitle

\ifCLASSOPTIONcompsoc
\IEEEraisesectionheading{\section{Introduction}\label{sec:introduction}}
\else
\section{Introduction}
\label{sec:introduction}
\fi

\IEEEPARstart{F}{raud} steals 80 billion dollars a year across all lines of insurance.
Fraud includes about 10 percent of property-casualty insurance losses and loss adjustment expenses each year; and
Property-casualty fraud thus equals about 34 billion dollars each year.
Data from ACI Worldwide, 14 out of 17 countries studied indicated an increase in fraud. However, the U.S. was hit the hardest of all, with almost half of all consumers have been a victim of fraud.
This made America the third-highest rate of fraud worldwide and was the only country to stand in the top three from 2014 to 2016. All these facts demonstrate the importance of fraud detection.\\
Fraud detection is a vital topic that applies to many industries including the financial sectors, banking, government agencies, insurance, and law enforcement, and more.  Fraud endeavors have detected a radical rise in current years, creating this topic more critical than ever. Despite struggles on the part of the troubled organizations, hundreds of millions of dollars are wasted to fraud each year.  Because nearly a few samples confirm fraud in a vast community, locating these can be complex.\\
Data mining and statistics help to predict and immediately distinguish fraud and take immediate action to minimize costs. Using developed data mining tools such as machine learning, association rules,  decision trees (Boosting trees, Classification trees, CHAID, and Random Forests), cluster analysis, and neural networks, imminent patterns can be produced to measure information such as the possibility of a fraudulent act or the dollar measure of fraud.  These anticipating models can help to focus resources most efficiently to prevent or recuperate fraud losses.\\
In this paper, we applied one class classifiers for fraud detection. We implemented SVDD as one of the famous one class classifiers. SVDD first was introduced by \emph{David Tax and Robert Doin}[19]. It is a kind of one-class classification methods based on SVM. By including the objective data inside a minimum hyper-sphere, a boundary around the objective data is constructed by SVDD. Inspired by SVMs, the SVDD determination boundary is explained by a few target objects, known as support vectors[9].\\
Our aim to use one class classifiers is that we need fewer data, and acquiring fraud data is a complicated task. Then we compared our method results with two class classifiers such as SVM. Also, a new method was represented by us for selecting training data which significantly improved the results.\\
Main contributions of this paper are applying one class classifiers methods for fraud detection and introducing an extension of \emph{DBSCAN} algorithm which is named \emph{REDBSCAN} for reducing numbers of samples. \\
Our paper is coordinated as follows. Section 2 defines previous attempts and struggles toward fraud detection. Section 3 represents our proposed method. Section4 presents results and compares them. Section5 concludes the paper and summarizes the achieved objects and perspectives.

\section{Related Work}

\IEEEPARstart EWT Nagi and collaborators mentioned different approaches toward fraud detection. First, we will take a quick look at data-mining techniques, and then other methods.\\
\emph{SVM} is a supervised machine learning method that is mostly used for binary classification. \emph{Houssem Eddine Bordjiba} and collaborators used SVM as their classifier[4]. Also, it is trained for Fraud Detection in Mobile Telecommunication Networks by \emph{Sharmila Subudhia and Suvasini Panigrahib}[10].\\
SVM is also used in other articles such as Nontechnical Loss Detection for Metered Customers
in Power Utility[12], Detecting Fraud of Credit Card  by applying
Questionnaire-Responded Transaction Model which is based on Support Vector Machines[13], Network-Based Intrusion Detection[14], and for detecting top management fraud[15].\\
A full review of the research which was conducted between 1997 and 2008 on using data mining methods for detecting financial fraud is presented by Yeh and Lien(2009)[27].\\
A combination of SVDD and Genetic algorithm is used for fault detection[19].
\emph{Christine Hines} and coauthor applied Machine Learning techniques for the first time in the restaurant industry to detect insider fraud in the point-of-sales transaction. They employed SVM and Random-Forest[29].
\\ 
RBML is the recognition and utilization of a set of relational rules that collectively express the information obtained by the system.
\emph{Ankit Kumar Jain, B. B. Gupta }implemented the rule-based data mining classification method in the discovery of \emph{smishing} messages and they reached higher than 99 percent true negative rate[5].\\
 OCSVM produces a border that includes the normal data points, and any external point would be abnormal. In the paper \emph{ to analyze the functional and physical } for detection of anomalies in press-hardening, they implemented \emph{OCSVM}, \emph{ANN}, and \emph{Isolation Forest (IF)} algorithms.
 They then compared these algorithms in terms of implementation, training time, execution time, precision, recall, and accuracy. As a result, ANN has an extended training time and a more complex development; however, it achieves a higher precision. The prediction time for any of the processes is quick, and all of them are well satisfied to be employed in near real-time[5].\\
Shing-Han Li and coauthors applied the Bayesian Classification and Association Rule to recognize the symbols of fraudulent accounts and the patterns of fraudulent transactions. They generated the detection rules based on the identified notes and employed to the layout of a fraudulent account detection system[22].\\
 Regression is a statistical methodology that is employed to expose the connection within one or more independent variables and a dependent variable (that is continuous-valued) [24]. Various practical examinations have employed logistic regression as a benchmark. The regression method is typically initiated using such mathematical rules as logistic regression and linear regression, and it is applied in the discovery of the frauds of the crop and automobile insurance, credit card, and fraud in corporations[7].\\
 \emph{CCM}  which is a cluster based classification technique executes classification by first putting data points based on the clear features in groups and, then SVM is utilized to distinguish the examples in each cluster. \emph{Sarwat Nizamani } and collaborators used this approach. They also implemented their task with SVM, decision tree and Naive Bayes and compared them in terms of results[8].\\
 k-Means clustering is a cluster examination algorithm in which users define k clusters that are not joint to each other by the feature price of the things to be classified. They proposed a Network Data Mining procedure that expands the K-mean clustering algorithm with a purpose to isolate periods with normal and abnormal traffic in the training dataset. In this approach then the centroids of the resulting cluster are applied for quick anomaly discovery in the monitoring of fresh data [7].\\
k-Medoids and k-Means algorithm are very similar to each other. Their fundamental difference is in the k-Medoids description of the various clusters. In it, the most centric object in the cluster determines every cluster and the implicit mean has no part since it may not refer to the cluster. The k-medoids approach is more robust than the k-means algorithm in the presence of noise and outliers since a medoid is less affected by outliers or other extreme values than a mean. The k-medoids identifies those network anomalies which comprise unknown intrusion. In terms of accuracy, It produces much better results than kMeans, and It has been compared with various other clustering algorithms[7].
  A branch of kMeans which designates an object to the cluster to that it fits, based on the mean of the cluster is called EM Clustering.
 In this method, there are no stringent limits between the clusters. In other words, instead of allocating an object in the dedicated cluster, assign the object to a cluster in order with a weight depicting the probability of association. In this extension new mean is computed by weight measures. In comparison to k means and k medoids, EM beat them and acquired greater accuracy [7].\\ 
 GA is employed to collect a set of classification rules from the network audit data in intrusion detection. The support-confidence framework is employed as a fitness function to adjudicate the quality of each rule. 
GA owns essential features such as robustness against noise and capacities to learn. In the case of anomaly detection, the benefits of GA methods proclaimed are high attack discovery rate and lower false-positive rate [7].\\
Also, Neural Network has been employed in fraud detection. Graham Williams and collaborators have implemented RNN for anomaly detection in data mining[20]. Another work which took advantage of NN is \emph{CARDWATCH}. It is a Neural Network Based Database Mining system for detecting fraud in Credit Cards.   This scheme is based on a NN learning module, presents an interface to a diversity of marketing databases, and its graphical user interface is straightforward. Test results received for synthetically produced credit card data and an auto-associative neural network model confirm very prosperous fraud detection rates[21].\\
Another NN is used for racing up the data mining and knowledge discovery process for detecting the fraud of Credit Card. This approach demonstrates the importance of the combination of neural networks and data mining [23].\\
A new approach such as a combination of Neural Network and data-mining called \emph{Neural Data Mining for Credit Card Fraud Detection} was presented by \emph{R. Brause} and collaborators[25]. This article explains how successfully neural network can be joined with high-level data mining methods to reach a high fraud coverage linked with a low false alarm rate.\\ 
  Hybrid approaches are those that can be performed by either connecting or joining different algorithms such as Combining supervised and unsupervised methods and Cascading supervised techniques.\\
The efficiency of the anomaly detection rate can be extremely enhanced by using an unsupervised algorithm as the performance of the supervised algorithm can hugely be risen in Combining supervised and unsupervised techniques. 
 Merging entropy of network features and SVM have been proposed that exceeded both SVM and entropy methods independently. Also, kMeans and ID3 algorithms were combined to classify anomalous and normal actions in computer Address Resolution Protocol (ARP) traffic[7].\\
   A new hybrid method which combines the SVM with and Multi-Verse Features Extraction (MVFEX) was proposed by \emph{Ali Safa Sadiq} and collaborators[28].\\
  Different classification algorithms are combined in order to achieve better accuracy which is called cascading supervised techniques. A mixture of naive Bayes and decision tree algorithm was recommended which focused on the construction of the achievement of Naive Bayesian (NB) classifier and ID3 algorithm. In Knowledge Data Discovery (KDD) cup dataset this hybrid algorithm was certified and accomplished 99 percent accuracy. 
Decision Tree and SVM were merged to define the ensemble approach that used Support Vector Machine (SVM), Decision Tree (DT) and hybrid DT-SVM classifier using waits. On test dataset, the ensemble approach produced 100 percent accuracy. Several strategies can be suggested since numerous types of aggregates are plausible. Also, the best resulting strategies can be executed practically[7].\\
 New hybrid approaches such as the combination of Genetic algorithm and SVM is presented by \emph{J.Nagi} and collaborators for Detection of Abnormalities and Electricity Theft[11]. Also, GA based on Fuzzy C-Means clustering and several supervised classifiers were merged as a model for automobile insurance fraud detection[16]. Also, a mixture of fuzzy rules and the Genetic Algorithm was introduced for detecting online fraud[26].\\
\section{Proposed Method}
Obtaining fraud samples for the training phase is one of the big challenges in this research area. So, in this article, to deal with this challenge we applied SVDD as one of the famous one class classifiers for fraud detection.\\ Advantages of using one class classifiers are that we just need one class data for training, and there is no need for acquiring fraud data which is a complex task. We compared our results with two class classifiers such as SVM.\\
Also, we presented a new approach for selecting samples since we are dealing with a huge number of data.\\
Our proposed method for fraud detection is illustrated in the below flowchart:\\
\\
\begin{figure}[h!]
\includegraphics{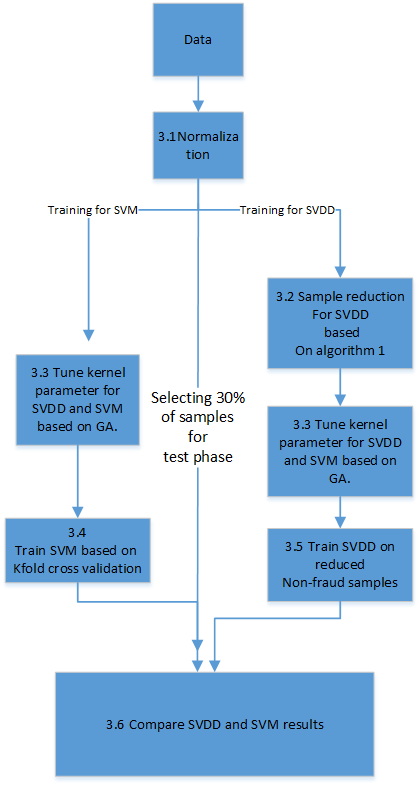}
\caption{Structure of the paper}
\end{figure}

\subsection{Normalization}
Dividing data set to fraud and non-fraud is essential since SVDD is one classifier algorithm and we could train it based on only non-fraud ones. Also, bank data may have large digits so, we need to normalize them by dividing every column to the maximum number in that column as a result, every data will be between 0 and 1. The purpose of this action is to process them easier.\\
Due to the availability of non-fraud data over fraud data and difficult process for acquiring the fraud data, non- fraud data is selected for the training phase for the SVDD algorithm and this is one of the advantages of using SVDD in fraud detection. We can train our classifier based on only non-fraud samples.\\
\subsection{Sample reduction for SVDD based on algorithm 1}
Due to the huge volume of data, the training phase for SVDD will be prolonged since we have one million data in our data set. SVDD is a kind of algorithm which is based on boundaries. As a result data's shape is important and should be kept.\\
We purposed our algorithm for this process which we will explain.
Our strategy for choosing training data instead of selecting randomly is to focus on data which will hold the general shape of data, and a few numbers of them will be used for the training phase. To implement this aim, we used \emph{DBSCAN} and an extension of it which is presented by us. This algorithm is named \emph{REDBSCAN}.
In \emph{DBSCAN} algorithm we consider every row that contains our data, a node. We assign weight to these nodes based on distance called \emph{K} that we calculate it. This process follows as for every node it's neighbors in range of k define the node's weight. Nodes with the highest weights are chosen. Now we can remove these node's neighbors since chosen nodes can keep the data's shape. As a result, this method provides lesser than 0.1 of non-fraud data and a faster training phase for SVDD.\\
We will illustrate the algorithm 1 with an example.\\
\\
\begin{figure}[h!]
\includegraphics[width=0.5\textwidth]{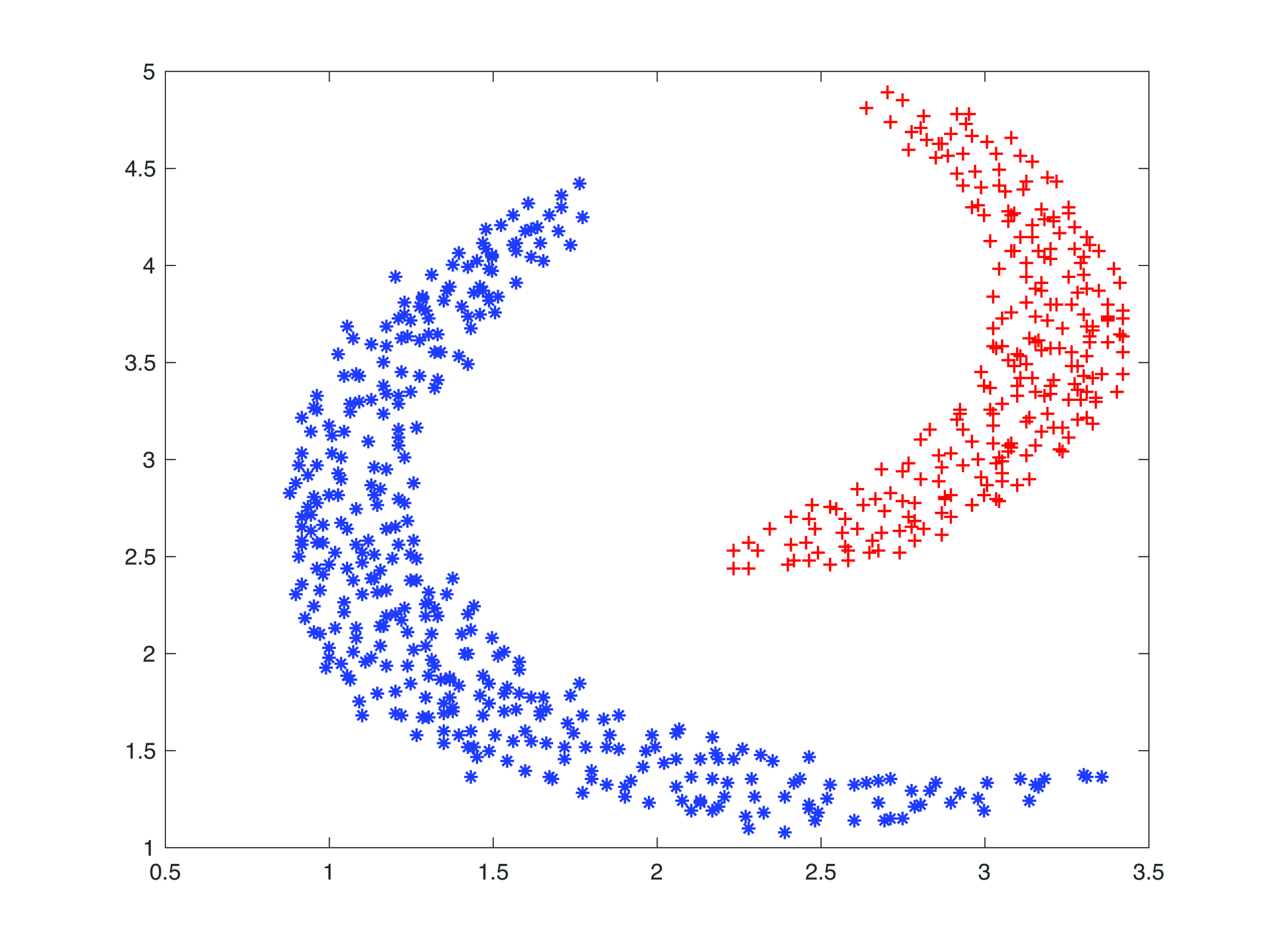}
\caption{Data of two classes}
\end{figure}
\\
In the figure 2, data for two classes can be distinguished, and it is clear that they hold a  special shape with a specific boundary. Now we desire to implement our algorithm with the aim of reducing training data without any changes in the data's shape.\\ 
After applying our proposed method the REDBSCAN algorithm, we select samples from data and in the figure 3, we can see the results.\\
This picture indicates that by implementing our algorithm numbers of training data for boundary algorithms such as SVDD will be reduced remarkably without changing the data's shape and our results.\\
\begin{figure}[h!]
\includegraphics[width=0.5\textwidth]{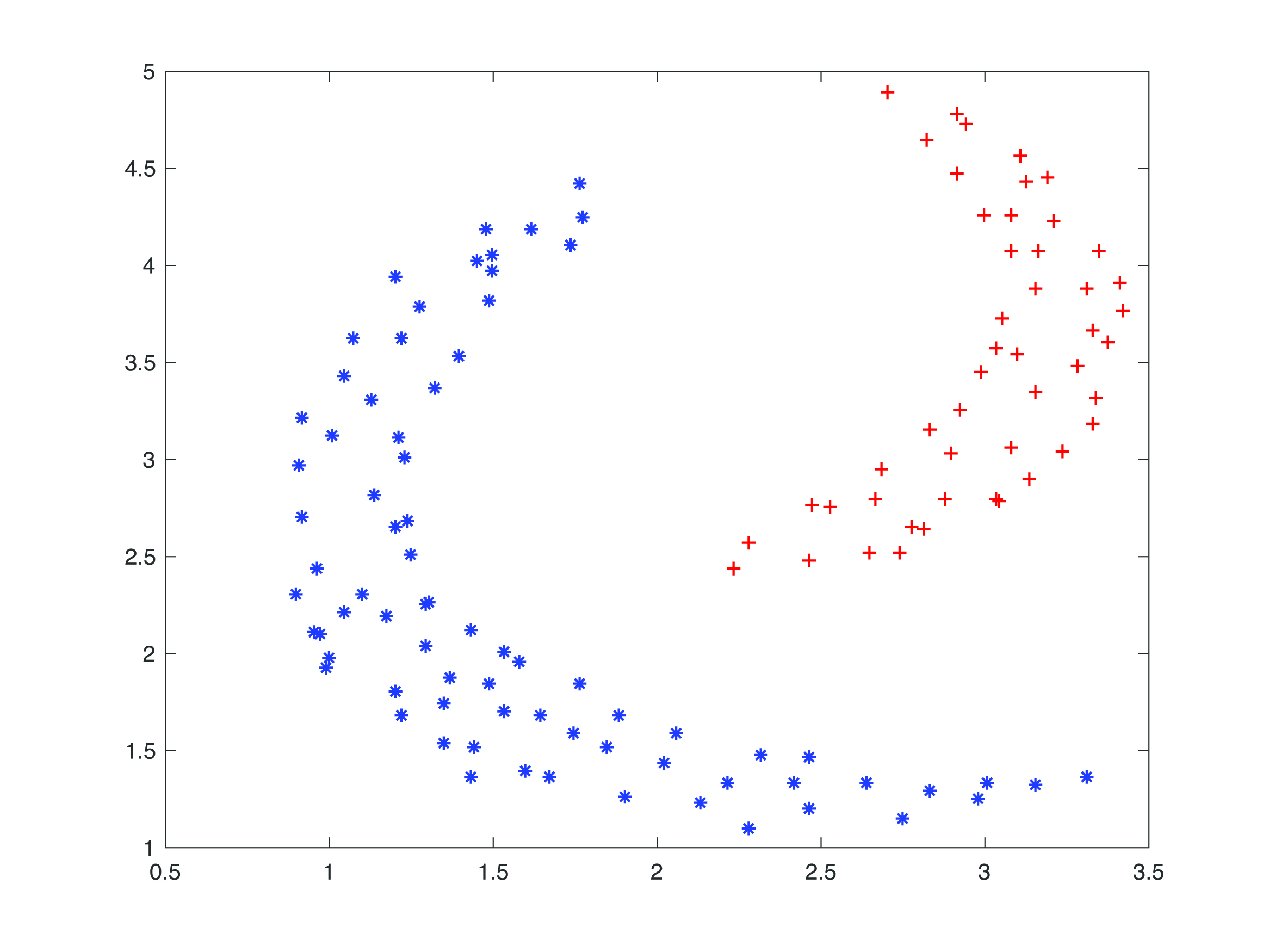}\\
\caption{Data of two classes after applying the REDBSCAN}
\end{figure}
\begin{algorithm}
\SetAlgoLined
\KwResult{Selected Samples}
 input: learning data, DBData, minpts, Eps, dim\; 
  M,N = Size(LearnData) \;
 \If {InputArg $ \leq 3$ \quad or \quad isempty(Eps)}{
   Eps = epsilon(LearnData,DBData)\;
}
 LearnData = [[1:m]' LearnData] \;
   DBData =  [[1:m]' DBData] \;  
 \While{isempty(LearnData) != null}{
  ob = x(1,:)\;
  D = dist(ob(2:n),x(:,2:n)) \; 
  i1 = find(D $\leq$ Eps) \;  
  max-in-subcluster = i1(find(y(i1,dim+6) == max(y(i1,dim+6))))\; 
  Selected-Samples = [Selected-Samples ; y(max-in-subcluster,2:dim+6)]\; 
  x(i1, :) = [ ] \;  
  y(i1, :) = [ ] \; 
 }
 \caption{REDBSCAN algorithm}
\end{algorithm}
This algorithm is based on DBSCAN and first constructs the DBdata based on it.
For calculating the radios for selected samples, Euclidean Distance weight function is used. Then best node with the most weight within the selected radios are chosen and their label is set.      
\subsection{Tune kernel parameter for SVDD and SVM based on GA}
In addition to learning data, SVDD requires parameters such as Sigma for RBF kernel and FARCREJ which is an error on the target class. These parameters values are crucial in the learning phase and there is no specific algorithm for tunning them.\\   
For tunning SVDD parameters, we benefited from the Genetic algorithm. Different methods could be used such as Neural Network. After this phase, we will acquire the best parameters for training data in SVDD.

\subsection{Train SVM based on k-fold cross validation}
 Cross-validation is a resampling strategy used to appraise machine learning models on a limited data sample. The number of combinations that a given data pattern is to be split into is called k which is a single parameter of the procedure. In applied machine learning, cross-validation is mainly used to predict the ability of models on unseen and new data. That is, to use a limited example with a purpose to predict how the model is assumed to act in general when employed to make forecasts on data not utilized during the training phase. We used this method while training SVM on our dataset.\\
The overall procedure is as follows:\\
1. Randomly the dataset is shuffled.\\
2. Dataset is split into K groups.\\
3. For each novel group:\\
4. We use the group as a test data set  or a holdout, and we use the remaining groups as a training data set.\\
5. We will try to fit a pattern on the training set and appraise it on the test data set.\\
6. After that, we retain the evaluation score and abandon the model. Then we
review the ability of the pattern by applying the sample of pattern evaluation rates.
Above all, each observation in the data sample is assigned to an individual group and for the duration of the procedure remains in that group. This points out that each sample is allowed to be used in the hold out set one time and used to train the model k-1 times.

\subsection{Train SVDD on reduced Non-fraud samples}
After applying DBSCAN and our proposed algorithm REDBSCAN, samples are reduced. These data are the best for training SVDD because SVDD is a boundary algorithm and the shape of data should be kept. Now we train SVDD on these reduced Non-fraud samples which will lead to faster computing and using efficiently of memory.

\subsection{Comparing SVDD and SVM results by testing samples}
In this section, we chose 30 percent of the whole dataset including fraud and non-fraud data accidentally in the test phase both for SVDD and SVM. According to previous sections we used just 10 percent of non-fraud data for training SVDD but,  SVM possesses 70 percent of both data(fraud and non-fraud) for the same process. Results based on the \emph{AUC} is interesting, for SVDD it is 0.9775 and SVM is 0.9460.\\
SVDD has acquired lesser than 10 percent of non- fraud data( about 1 million) in training chapter and all these data are from one training class(non-fraud), this method is suitable since obtaining non-fraud data is much easier and the need for other classes will be eliminated. On the other hand, in SVM huge amount of training data that is 70 percent of all dataset (about 700800) will make the training development long and expensive also, demands to access to both classes, but, obtaining fraud data is tough.
All of the above points are the disadvantages of SVM and demonstrates that SVDD outperforms the SVM.\\
For a better comparison between the two algorithms, we decrease the number of training data for SVM and arrange them as same as SVDD. As it was expected, \emph{AUC} for SVM is declined but, in SVDD it is still the same number. This point demonstrates the advantage of using the SVDD algorithm and one classifier over other approaches.  
\section{Experiments}
We implemented algorithms such as SVM, SVDD, and compared them in terms of precision, recall, AUC, and FM(f-measure).
\subsection{Data-set}
We use the \emph{Paysim synthetic dataset of mobile money} dataset.  It is balanced down 1/4 of the primary dataset and was exhibited in the paper named \emph{ PaySim: A financial mobile money simulator for fraud detection}. This dataset consists of approximately one million data and about 1142 are labeled fraud [17] \\
Due to the huge volume of information in the data set, we decided to divide it into two files in \emph{xlsx}formation to ease the process of reading the data. To implement the SVDD algorithm, we took advantage of Matlab toolboxes such as dd-tools and pr-tools that are presented by \emph{D.M.J. Tax} and collaborators.
\subsection{Results}
In this section, results for SVDD and SVM are illustrated in numbers and figures.\\
In this table, we compare the one class classifier algorithm(SVDD) and SVM:
\begin{center}
\begin{tabular}{ |c|c|c| }
 \hline
 Comparison method & SVDD & SVM \\ 
 \hline
 AUC & 0.9775 & 0.9460 \\ 
 \hline
 precision & 0.9194 & 0.8441 \\ 
 \hline
 recall & 0.8557 & 0.7550 \\ 
 \hline
 f-measure & 0.8864 & 0.7971 \\ 
 \hline
\end{tabular}
\end{center}
The figures below demonstrates and compares the ROC for SVDD and SVM algorithms:\\
ROC for SVDD and SVM:\\

\begin{figure}[h!]
\includegraphics[width=0.5\textwidth]{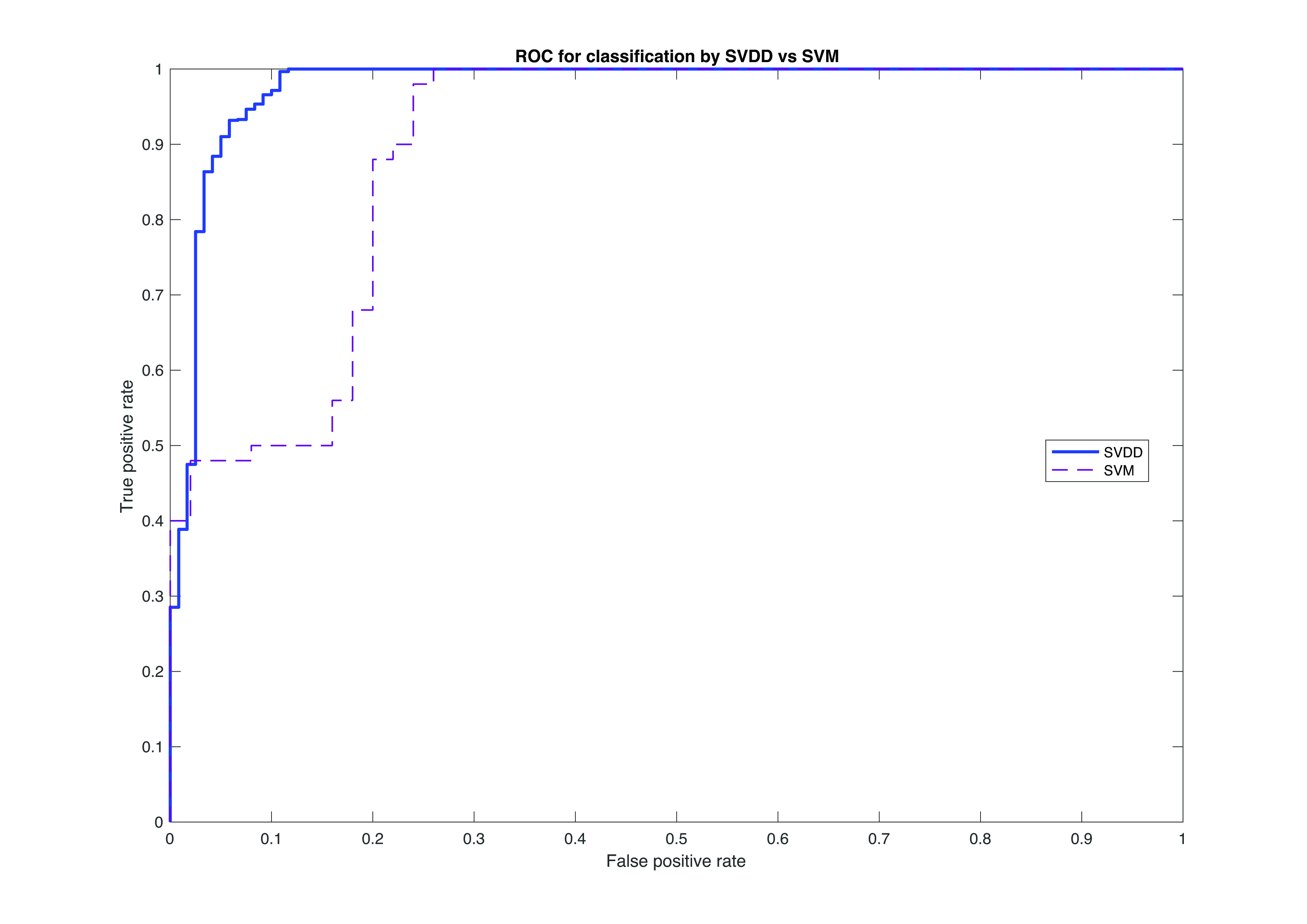}
\caption{ROC of SVDD and SVM}
\end{figure}

In case of speed, when we applied the SVDD algorithm without REDBSCAN, the training phase takes about 194 seconds. But, when REDBSCAN is applied to the dataset, the training phase takes about 1.69 second which is remarkably faster.\\
REDBSCAN reduces the samples and chooses the proper ones, which keep the shape of data so, the training phase is faster.

\section{Conclusion}
A novel approach for fraud detection is proposed in this paper. We applied one class classification methods for detecting fraud. These techniques do not require the fraud sample for the training phase. Since acquiring fraud samples is hard and complex, this can be regarded as one of the advantages of our proposed method. In the training phase, we are dealing with a huge number of data so, defining an approach for choosing the proper samples which can keep the shape of data and lead to faster learning. Our proposed method named REDBSCAN is an extension of the DBSCAN algorithm and achieved the desired results faster.\\
The results showed that implementing the proposed method improved the fraud detection process in comparison with two class classifiers in both performance and speed.
\section{Future work}
In the future, we will develop an approach for tuning the SVDD parameters to achieve better results.


%





\begin{thebibliography}{1}

\bibitem{IEEEhowto:kopka}
EWT Ngai, Y Hu, YH Wong, Y Chen, X Sun, \emph{The application of data mining techniques in financial fraud detection: A classification framework and an academic review of literature
}, \relax Decision Support Systems, 2011.
\bibitem{IEEEhowto:kopka}
 ShiguoWang, \emph{A Comprehensive Survey of Data Mining-Based Accounting-Fraud Detection 
}, \relax International Conference on Intelligent Computation Technology and Automation, 2010.

\bibitem{IEEEhowto:kopka}
 Anuj Sharma, Prabin Kumar Panigrahi  \emph{A Review of Financial Accounting Fraud Detection based on Data Mining Techniques 
}, \relax International Journal of Computer Applications, February 2012.

\bibitem{IEEEhowto:kopka}
 Houssem Eddine Bordjiba, ElMouatez Billah Karbab, Mourad Debbabi  \emph{Data-driven approach for automatic telephony threat analysis and campaign detection 
}, \relax DFRWS 2018 Europe d Proceedings of the Fifth Annual DFRWS Europe.

\bibitem{IEEEhowto:kopka}
Ankit Kumar Jain, B. B. Gupta \emph{Rule-Based Framework for Detection of Smishing Messages in Mobile Environment 
}, \relax 6th International Conference on Smart Computing and Communications, ICSCC 2017.

\bibitem{IEEEhowto:kopka}
Erik Lejona, Petter Kyosti, John Lindstrom \emph{Machine learning for detection of anomalies in press-hardening: Selection of efficient methods
}, \relax 51st CIRP Conference on Manufacturing Systems.

\bibitem{IEEEhowto:kopka}
Shikha Agrawal, Jitendra Agrawal \emph{Survey on Anomaly Detection using Data Mining Techniques 
}, \relax Egyptian Informatics Journal (2014).


\bibitem{IEEEhowto:kopka}
Sarwat Nizamani, Nasrullah Memon, Mathies Glasdam, Dong Duong Nguyen  \emph{Detection of fraudulent emails by employing	 advanced feature abundance
}, \relax 19th International Conference on Knowledge Based and Intelligent Information and Engineering Systems.


\bibitem{IEEEhowto:kopka}
Mohammad GhasemiGol,  Mostafa Sabzekar, Reza Monsefi, Mahmoud Naghibzadeh, Hadi Sadoghi Yazdi  \emph{A New Support Vector Data Description with Fuzzy Constraints 
}, \relax 2010  International Conference on Intelligent Systems, Modelling and Simulation.

\bibitem{IEEEhowto:kopka}
Sharmila Subudhia, Suvasini Panigrahib  \emph{Quarter-Sphere Support Vector Machine for Fraud Detection in Mobile Telecommunication Networks
}, \relax International Conference on Intelligent Computing, Communication Convergence  (ICCC-2014).

\bibitem{IEEEhowto:kopka}
J. Nagi, K. S. Yap, S. K. Tiong, Member, IEEE, S. K. Ahmed, Member, IEEE, A. M. Mohammad  \emph{Detection of Abnormalities and Electricity Theft using Genetic Support Vector Machines
}, \relax IEEE.

\bibitem{IEEEhowto:kopka}
Jawad Nagi, Keem Siah Yap, Sieh Kiong Tiong, Member, IEEE, Syed Khaleel Ahmed, Member, IEEE, and Malik Mohamad  \emph{Nontechnical Loss Detection for Metered Customers in Power Utility Using Support Vector Machines
}, \relax IEEE TRANSACTIONS ON POWER DELIVERY, VOL. 25, NO. 2, APRIL 2010.

\bibitem{IEEEhowto:kopka}
Rong-Chang Chen1, Ming-Li Chiu2, Ya-Li Huang2, and Lin-Ti Chen2  \emph{Detecting Credit Card Fraud by Using
Questionnaire-Responded Transaction Model Based on Support Vector Machines
}

\bibitem{IEEEhowto:kopka}
Dong Seong Kim and Jong Sou Park  \emph{Network-Based Intrusion Detection with Support Vector Machines
}

\bibitem{IEEEhowto:kopka}
Ping-Feng Pai, Ming-Fu Hsu, Ming-Chieh Wang  \emph{A support vector machine-based model for detecting top management fraud
}, \relax journal homepage:www.elsevier.com/locate/knosys

\bibitem{IEEEhowto:kopka}
Sharmila Subudhi, Suvasini Panigrahi  \emph{Use of optimized Fuzzy C-Means clustering and supervised classifiers for automobile insurance fraud detection
}, \relax Journal of King Saud University Computer and Information Sciences
\bibitem{IEEEhowto:kopka}
Edgar Alonso Lopez-Rojas, Ahmad Elmir, andStefan Axelsson
  \emph{PAYSIM: A FINANCIAL MOBILE MONEY SIMULATOR FOR FRAUD DETECTION
}, \relax Conference: 28th European Modeling and Simulation Symposium 2016 (EMSS 2016)At: Larnaca, Cyprus

\bibitem{IEEEhowto:kopka}
Tao Xin-min, Chen Wan-Hai, Du Bao-Xiang, XuYoung, Dong Han-Guang
  \emph{PAYSIM: A Novel Model of one-class Bearing Fault Detection using SVDD and Genetic Alghorithm
}, \relax 2007 Second IEEE Conference on Industrial Electronics and Applications

\bibitem{IEEEhowto:kopka}
David M.J. Tax, Robert P.W. Duin
  \emph{Support vector domain description
}, \relax Pattern Recognition Letters 20, pp. 1191-1199, 1999. 

\bibitem{IEEEhowto:kopka}
Graham Williams, Rohan Baxter, Hongxing He, Simon Hawkins 
  \emph{A Comparative Study of R" for Outlier Detection in Data Mining
}, \relax 2002 IEEE.

\bibitem{IEEEhowto:kopka}
Emin Aleskerov, Bernd fieisleben and Bharat Rao
  \emph{CARDWATCHA: A Neural Network Based Database Mining system for Credit Card Fraud
}, \relax 

\bibitem{IEEEhowto:kopka}
Shing-Han Li, David C. Yen, Wen-Hui Lu, Chiang Wanga
  \emph{Identifying the signs of fraudulent accounts using data mining techniques
}, \relax Computers in Human Behavior(2018) 

\bibitem{IEEEhowto:kopka}
Mubeena Syeda, Yan-Qing Zbang and Yi Pan 
  \emph{Parallel Granular Neural Networks for Fast Credit Card Fraud Detection
}, \relax 2002 IEEE

\bibitem{IEEEhowto:kopka}
J. Han, M. Kamber, 
  \emph{ Data Mining: Concepts and Techniques
}, \relax Kaufmann Publishers, 2006, pp. 285,464.

\bibitem{IEEEhowto:kopka}
R. Brause, T. Langsdorf, M. Hepp
  \emph{Neural Data Mining for Credit Card Fraud Detection
}, \relax 

\bibitem{IEEEhowto:kopka}
Mohammad Reza Parsaei, Reza Javidan, Mohammad Javad Sobouti
  \emph{Optimization of Fuzzy Rules for Online Fraud Detection with the Use of Developed Genetic Algorithm and Fuzzy Operators
}, \relax Asian Journal of Information Technology 2016.

\bibitem{IEEEhowto:kopka}
Yeh, I.C and C.H Lien
  \emph{The comparisons of data mining techniques for the predictive accuracy
  of probability of default of credit card clients. 
}, \relax Exp. Syst. Appli. 2009

\bibitem{IEEEhowto:kopka}
Ali Safa Sadiq, Hossam Faris, Ala M. Al-Zoubi, Seyedali Mirjalili,
and Kayhan Zrar Ghafoor
  \emph{Fraud Detection Model Based on Multi-Verse
Features Extraction Approach for Smart City
Applications. 
}, \relax Smart Cities Cybersecurity and Privacy
2019

\bibitem{IEEEhowto:kopka}
Christine Hines, Abdou Youssef
  \emph{Machine Learning Applied to Rotating Check Fraud Detection. 
}, \relax 2018 1st International Conference on Data Intelligence and Security (ICDIS)


\end{thebibliography}
\end{document}